\documentclass[letterpaper, 10 pt, journal, twoside]{ieeetran}
\usepackage{amsmath,amsfonts}
\usepackage{algorithmic}
\usepackage{algorithm}
\usepackage{array}
\usepackage[caption=false,font=normalsize,labelfont=sf,textfont=sf]{subfig}
\usepackage{textcomp}
\usepackage{stfloats}
\usepackage{url}
\usepackage{verbatim}
\usepackage{graphicx}
\usepackage{cite}
\usepackage{booktabs} 
\usepackage{amsmath}
\usepackage{multirow}
\usepackage[hidelinks]{hyperref}
\usepackage[capitalise]{cleveref}
\crefname{figure}{Fig.}{Figs.}
\Crefname{figure}{Fig.}{Figs.}
\crefname{section}{Section}{Sections}
\Crefname{section}{Section}{Sections}
\crefname{table}{Table}{Tables}
\Crefname{table}{Table}{Tables}

\newcommand{\std}[1]{{\normalfont\tiny$\pm$#1}}
\newcommand{\meanstd}[2]{%
    \mbox{#1\std{#2}}%
}
\newcommand{\bestmeanstd}[2]{%
    \mbox{\textbf{#1}\std{#2}}%
}
\newcommand{\secondmeanstd}[2]{%
    \mbox{\underline{#1}\std{#2}}%
}

\begin{document}

\title{Fetch My Beer: Synthetic-to-real Hierarchical Policy for Smooth Pick-and-place}

\author{Yingyue Li$^{1,*}$,
Chenyangguang Zhang$^{2,*}$,
Ruida Zhang$^{1,*}$,
Bowen Fu$^{3}$,
Guangyao Zhai$^{4}$,
and Xiangyang Ji$^{1}$
\thanks{Manuscript received: June 26, 2026; Accepted August 19, 2026.}%Use only for final RAL version
\thanks{This paper was recommended for publication by Editor Markus Vincze upon evaluation of the Associate Editor and Reviewers’ comments.} %Use only for final RAL version
\thanks{$^{1}$Yingyue Li, Ruida Zhang and Xiangyang Ji are with Department of Automation, Tsinghua University, Beijing, China. {\tt\footnotesize \{yingyue-21, zhangrd23\}@mails.tsinghua.edu.cn, xyji@tsinghua.edu.cn}}%
\thanks{$^{2} $Chenyangguang Zhang is with ETH Zurich, Zurich, Switzerland. {\tt\footnotesize chenyangguang.zhang@inf.ethz.ch}}%
\thanks{$^{3} $Bowen Fu is with Beijing Xiaomi Robot Technology Co., Ltd, Beijing, China. {\tt\footnotesize fubowen@xiaomi.com}}%
\thanks{$^{4} $Guangyao Zhai is with Technical University of Munich, Munich, Germany. {\tt\footnotesize guangyao.zhai@tum.de}}%
\thanks{$^{*} $Equal contribution; Corresponding author: Xiangyang Ji.}
% $^\dagger$Corresponding author.}
\thanks{Digital Object Identifier (DOI): see top of this page.}
}
% Use only for final RAL version.

\markboth{IEEE Robotics and Automation Letters. Preprint Version. Accepted AUGUST, 2026}{Li \MakeLowercase{\textit{et al.}}: Fetch My Beer: Synthetic-to-real Hierarchical Policy for Smooth Pick-and-place}

% \IEEEpubid{0000--0000~\copyright~2026 IEEE}

\maketitle

\begin{abstract}
Many real-world robotic applications require dynamically sensitive manipulation, where success depends not only on reaching a target state but on maintaining stable object dynamics throughout execution. 
We study the stable transport of liquid-filled containers, where a robot must move objects to target locations while suppressing sloshing and preventing spillage. 
Unlike conventional pick-and-place, this task imposes stringent requirements on motion smoothness and trajectory-level stability, exposing clear limitations in existing systems. 
Specifically, fluid simulation remains too costly for online reinforcement learning; human teleoperation introduces unintended accelerations that induce sloshing during imitation learning; and current policy pipelines optimize for task completion rather than dynamic stability.
We propose a synthetic-to-real framework coupling physically validated data generation with a hierarchical, diffusion-based controller. 
The scalable data pipeline synthesizes grasps, filters unstable poses via a vision-language model, and validates transport trajectories through fluid simulation. 
The policy is organized with a high-level module that translates language and visual observations into SE(3) control targets, and a latent diffusion controller that first plans efficiently in a compact latent space and then decodes dense action chunks, enabling the high control frequency needed for smooth and stable motion.
Extensive experiments show our system outperforms state-of-the-art manipulation policies in transport smoothness and dynamic stability.
Our project page: \url{https://fetch-my-beer.github.io/}.
\end{abstract}


\section{INTRODUCTION}

\begin{figure}[t]
    \centering
    \includegraphics[width=\linewidth]{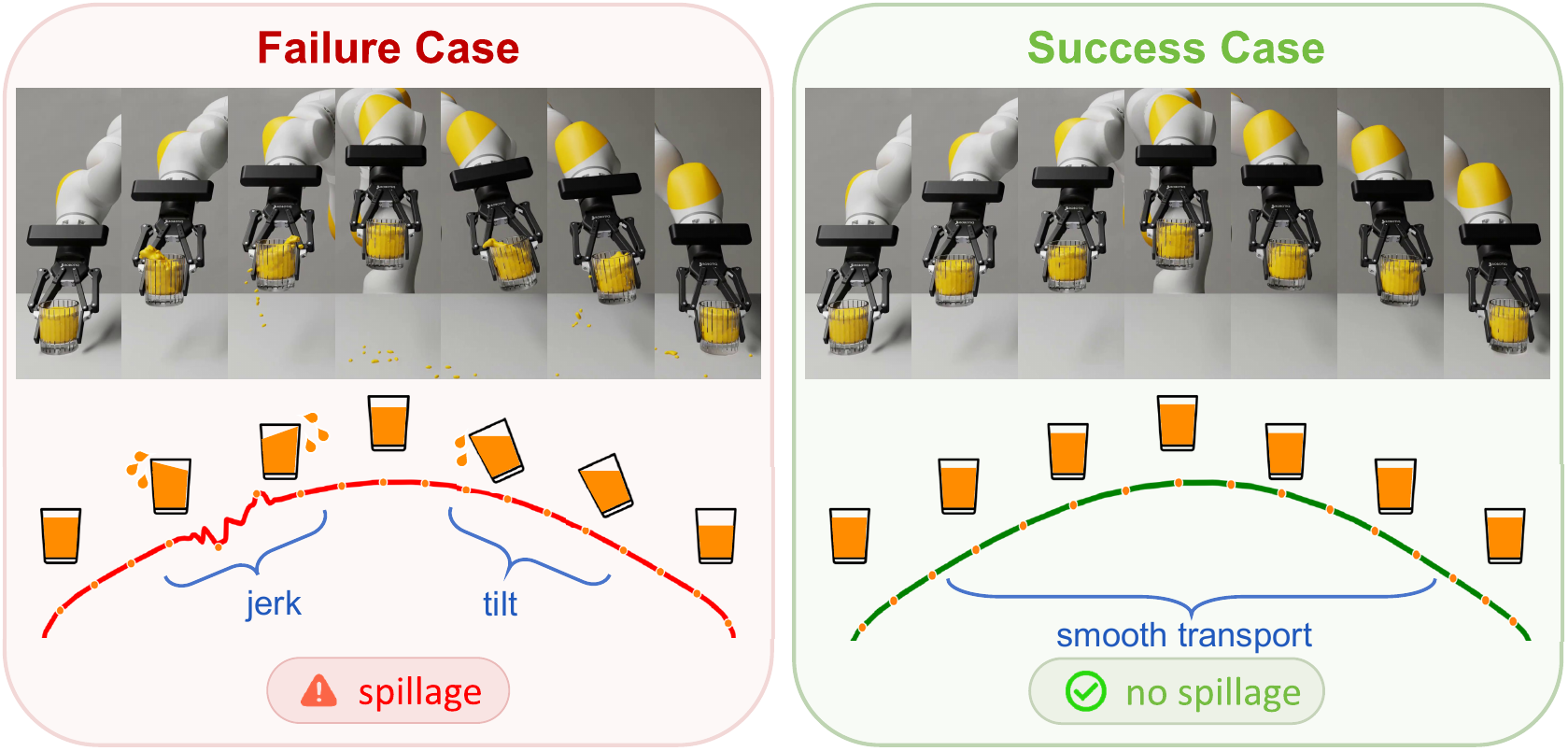}
    \caption{In dynamically sensitive tasks such as liquid transport, abrupt acceleration and excessive tilt lead to sloshing and spillage, whereas smooth, temporally consistent motion enables stable, spill-free transport.}
    \label{fig:teaser}
    \vskip -15pt
\end{figure}

\IEEEPARstart{R}{ecent} advances in robot manipulation~\cite{rvt2, act, diffusion_policy, openvla, pi0, gr00t, pi05} have enabled remarkable progress in tabletop manipulation such as pick-and-place, rearrangement, and tool use.
However, many real-world tasks require dynamically sensitive manipulation, where success depends not only on reaching a goal pose, but also on maintaining stable object dynamics throughout execution.
Representative examples include transporting liquid-filled containers, carrying fragile or deformable objects, and manipulating items with unstable contents, which is highly relevant to service robotics, laboratory automation, and industrial delivery scenarios.
In these tasks, even small undesired accelerations, jitter, or abrupt corrective motions can induce irreversible dynamic disturbances, leading to failure despite correct final positioning.
In this work, we study stable robotic transport of liquid-filled containers, where a robot must move objects to target locations while suppressing sloshing and preventing spillage.
Compared with conventional pick-and-place tasks, such dynamically sensitive manipulation imposes substantially stricter constraints on motion smoothness and stability, requiring joint reasoning over semantics, geometry, and fine-grained control (See \cref{fig:teaser}).

Despite its practical importance, stable and smooth robotic manipulation remains highly challenging for existing robot learning systems. 
For Reinforcement Learning (RL) methods, the computational expense of fluid simulation poses a severe bottleneck. Because computing fluid dynamics is significantly slower than rigid-body simulation, performing the millions of online, interactive steps required for RL convergence is computationally prohibitive. 
Conversely, Imitation Learning (IL) allows for offline training but faces a fundamental difficulty in data collection.  
Human teleoperation for dynamically sensitive tasks is highly error-prone, since even small unintended accelerations can induce severe sloshing or instability. 
Although simulation-based data generation provides a scalable alternative, existing pipelines~\cite{rlbench, robotwin, robotwin2, maniskill, maniskill2, maniskill3} primarily optimize for rigid-body task completion and geometric feasibility, while overlooking motion stability and smoothness.
On the policy side, recent diffusion-based visuomotor policies~\cite{diffusion_policy, reactive_diffusion_policy, graspldp, dp1, dp2} and Vision-Language-Action (VLA) models~\cite{openvla, pi0, gr00t, pi05} perform well on general manipulation benchmarks, but primarily optimize geometric accuracy rather than dynamic stability.
In particular, dynamically sensitive manipulation tasks require policies to jointly perform semantic reasoning, long-horizon motion planning, and high-frequency smooth control under stringent stability constraints. This multi-scale requirement remains difficult for existing flat, end-to-end architectures to balance. 
As a result, enabling reliable and stable manipulation requires not only physically grounded data generation pipelines, but also control architectures explicitly designed for temporally coherent and stability-aware manipulation.

To address these challenges, we propose a comprehensive synthetic-to-real framework that combines physically validated synthetic data generation with a hierarchical, diffusion-based control architecture for stable and smooth robotic manipulation.
On the data generation side, we develop a scalable pipeline in NVIDIA Isaac Sim~\cite{isaaclab} to automatically produce high-quality demonstrations across diverse scenes and container categories.
Specifically, we generate diverse grasp candidates using state-of-the-art grasp models and leverage a Vision-Language Model (VLM) to filter out physically infeasible or geometrically unstable poses. 
We then compute smooth trajectories using motion planning and rigorously validate them through fluid simulation. 
This ensures that the collected dataset satisfies both geometric feasibility and stringent trajectory-level stability constraints.
On the policy side, we introduce a hierarchical control framework that decouples semantic task reasoning from high-frequency motion generation. 
At the high level, a keyframe policy infers task-relevant SE(3) sub-goals from language instructions and multi-view RGB-D observations. Conditioned on these sub-goals, the low-level controller employs a latent diffusion architecture: compact latent action sequences are efficiently generated in a lower-dimensional space, then subsequently refined into dense, executable trajectories via a decoder. 
This hierarchical design promotes temporally coherent, smooth motion generation while maintaining high-frequency responsiveness.

To evaluate manipulation under fluid-induced dynamics, we introduce a simulation benchmark and metrics that prioritize trajectory-level stability over mere task success. 
Our benchmark assesses policy generalization across diverse object geometries, workspace layouts, and visual shifts. 
To quantify performance, we introduce metrics that jointly measure task success, trajectory smoothness, container orientation stability, and worst-case tilt. 
Experiments demonstrate that our method consistently outperforms strong hierarchical diffusion and VLA baselines across both in-domain and out-of-domain settings. 
Furthermore, trained solely on synthetic demonstrations, our policy achieves effective zero-shot sim-to-real transfer without any real-world fine-tuning. 

Our main contributions are summarized as follows:
\begin{itemize}
    \item We develop a scalable synthetic data generation pipeline with physics-based fluid validation, enabling efficient collection of dynamically stable manipulation demonstrations and reliable sim-to-real transfer.
    \item We introduce a simulation benchmark together with evaluation metrics that capture both task success and trajectory-level dynamic stability, enabling systematic evaluation under in-domain and out-of-domain settings.
    \item We design a hierarchical latent diffusion framework that decouples semantic reasoning from high-frequency motion generation, achieving stable and smooth manipulation in simulation and zero-shot real-world scenarios.
\end{itemize}
\section{RELATED WORK}

\subsection{Dynamically Sensitive Manipulation}
Manipulating dynamically sensitive objects is a long-standing problem in robotics, where task success depends on both geometric accuracy and stable physical evolution.
Early studies on liquid transport~\cite{sloshing1, sloshing2, sloshing3, sloshing4, sloshing5} primarily relied on model-based control, trajectory optimization, and vibration suppression techniques to reduce fluid oscillation.
While these methods demonstrate strong performance in carefully modeled settings, they often require accurate fluid dynamics priors, handcrafted controllers, or fixed robot configurations, limiting their generalization to unstructured settings.
This has motivated increasing interest in learning-based approaches for stability-aware manipulation.

\subsection{Visuomotor Policy and Hierarchical Robot Control}
Behavior cloning (BC) methods~\cite{rvt2, act, diffusion_policy} learn policies from expert demonstrations, enabling robots to directly map visual observations and language instructions to continuous actions.
Large VLA models~\cite{openvla, pi0, gr00t, pi05} further improve task generalization through vision-language pretraining.
Direct dense action prediction, however, can be computationally expensive and temporally inconsistent, which is problematic when small action errors disturb object dynamics.
Hierarchical control mitigates this issue by decomposing manipulation across temporal abstraction levels. 
Prior works have explored hierarchical learning through sub-tasks, sub-goals, and multi-stage policies~\cite{hierarchical1, hdp, hierarchical3, hierarchical4}, which improve long-horizon reasoning and simplify low-level action  generation. 
We follow this principle and use event-labeled keyframes for task-level reasoning and latent diffusion for smooth, high-frequency control.

\subsection{Synthetic Data Generation}
Large-scale demonstration data is essential for training robust visuomotor policies, yet real-world collection of dynamically stable manipulation trajectories is expensive and difficult to scale.
Recent robotic simulation frameworks and synthetic data generation benchmarks, such as RoboTwin~\cite{robotwin,robotwin2} and ManiSkill~\cite{maniskill, maniskill2,maniskill3}, have enabled scalable demonstration generation through scripted expert policies, dense-reward RL, or human teleoperation.
However, most pipelines emphasize rigid-body completion and collision-free motion rather than dynamic stability along the full trajectory.
Our pipeline additionally applies physics-based validation, retaining trajectories that are both geometrically feasible and dynamically stable.

\section{SIMULATION} \label{sim}

\begin{figure*}[t]
    \centering
    \includegraphics[width=0.9\linewidth]{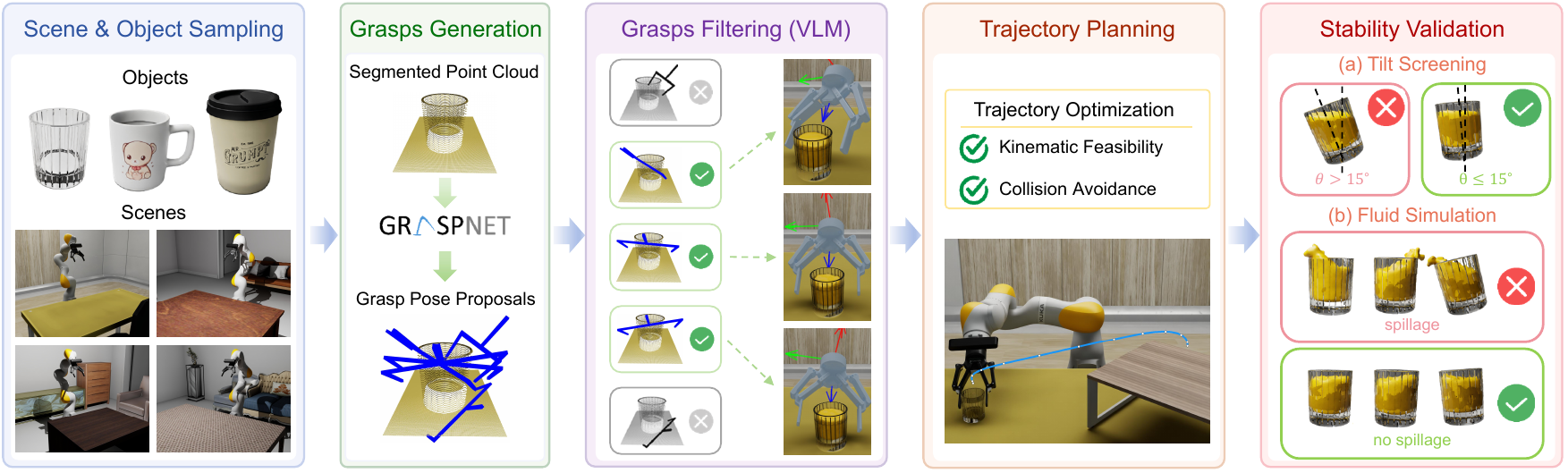}
    \caption{Stability-aware synthetic data generation pipeline. We sample diverse objects and scenes, generate 6-DoF grasp candidates from segmented point clouds, and use a vision-language model to score each rendered grasp proposal with a short feasibility reason. Selected grasps are passed to a smooth trajectory planner, which enforces kinematic feasibility and collision avoidance. Candidate trajectories are then filtered by tilt screening and physics-based fluid simulation to remove unstable or spill-prone motions.}
    \label{fig:sim_pipeline}
    \vskip -15pt
\end{figure*}
Training stable and precise manipulation policies requires a large volume of high-quality demonstrations that are kinematically feasible, smooth, and dynamically safe. 
Real-world teleoperation is expensive and unreliable for this purpose, as small control errors can easily destabilize the manipulated object.
While simulation offers a scalable alternative, existing pipelines primarily target rigid-body collision avoidance and goal reaching. 
For dynamically sensitive tasks, such geometric checks are insufficient: trajectories that reach the goal can still fail due to abrupt acceleration or invalid intermediate poses.
We therefore build a stability-aware synthetic data generation pipeline in NVIDIA Isaac Sim~\cite{isaaclab}, shown in \cref{fig:sim_pipeline}.
For each scene configuration, the pipeline proposes grasp candidates, filters implausible ones via a VLM, plans and optimizes grasp-to-place trajectories with cuRobo~\cite{curobo}, and validates them through rule-based tilt screening and physics-based fluid simulation. 
This pipeline enables scalable collection of high-quality demonstrations that satisfy semantic consistency, kinematic feasibility, and trajectory-level dynamic stability.

\subsection{Grasping Proposals}

We use AnyGrasp~\cite{anygrasp}, a state-of-the-art grasp pose estimation model, to generate diverse 6-DoF grasp hypotheses from rendered scene point clouds.
Given the target object geometry, the grasp model produces a ranked set of candidate gripper poses that covers multiple approach directions and contact regions.
Retaining multiple candidates is important for stable manipulation: a grasp that is reachable and force-closure feasible may still lead to poor downstream motion if it constrains the wrist, induces unfavorable object poses during transport, or leaves little margin for smooth placement.
We therefore keep a diverse candidate set rather than committing to the top geometric grasp directly.
This design allows the following stages to select grasps according to task-level plausibility and trajectory feasibility, not only local contact geometry.

\subsection{VLM-based Filtering}
Purely geometric grasp models do not account for whether a grasp is suitable for stable downstream motion, and their confidence scores may not reflect transport feasibility in a specific scene.
We therefore use Qwen3-VL~\cite{qwen3-vl} as a feasibility filter before motion planning.
Since the VLM operates on 2D inputs, we render each 6-DoF grasp by projecting a simplified gripper mesh onto the scene image, thereby explicitly visualizing the proposed gripper-object interaction.
Given this composite image, the VLM predicts a confidence score with a brief justification. 
This stage filters out candidates that are visually implausible or unsuitable for the task, such as grasps that collide with the table, insert the gripper into the container, approach the object from an infeasible direction, contact unstable regions, or result in an unfavorable object pose for subsequent transport. 
The remaining candidates are sorted by VLM confidence and passed to the motion planning stage. 
In this way, the pipeline combines the geometric coverage of grasp proposal networks with the visual reasoning and commonsense feasibility assessment of a VLM before costly trajectory execution and stability validation.

\subsection{Smooth Path Planning with Stability-aware Filtering}
To obtain high-quality robot end-effector trajectories that satisfy the stringent smoothness requirements for manipulating liquid-filled containers, we employ cuRobo~\cite{curobo}, a CUDA-accelerated library, as an efficient motion generation backend for inverse kinematics and trajectory optimization.
Given the grasp candidates filtered and ranked by VLM, we randomly sample multiple candidate placement poses on the target platform for each selected grasp and use cuRobo to solve the corresponding grasp-to-place motions. 
Among all these feasible motions, we select the trajectory with the best optimization quality, considering IK feasibility, collision avoidance, and trajectory cost.

To ensure that the collected data satisfy the strict smoothness and stability requirements of stable and smooth manipulation, we further apply two complementary filtering mechanisms: a rule-based screening criterion and physics-based fluid simulation.
The rule-based screening serves as a lightweight preliminary filter to quickly eliminate clearly unsuitable trajectories during execution.
Specifically, we monitor a dedicated Z-axis cup angle metric and reject trajectories in which the cup deviates from the vertical direction by more than an experimentally determined threshold of 15 degrees.
This criterion provides an efficient way to remove motions that are highly likely to cause spillage during transport.
Beyond the tilt-based geometric heuristic, we further evaluate the remaining trajectories with particle-based fluid simulation in Isaac Sim. 
This step is designed to capture spill-prone motions that cannot be detected by cup orientation alone. 
In particular, a trajectory may keep the container nearly upright while still inducing strong liquid sloshing due to abrupt velocity changes, high accelerations, or jerky motions during transport. 
Such dynamic effects are difficult to identify using rule-based tilt screening, but they are critical for stable liquid manipulation. 
We therefore use the Position Based Dynamics (PBD)~\cite{pbf} solver in Isaac Sim as a physics-based validation stage to reject trajectories that complete the geometric task yet cause excessive liquid oscillation or spillage.
Only trajectories that pass both filters are stored as demonstrations.
As a result, the dataset contains motions that are geometrically feasible, smooth at the trajectory level, and dynamically stable under the fluid-sensitive setting.

\section{Hierarchical Latent Diffusion Policy}

\begin{figure*}[t]
    \centering
    \includegraphics[width=0.75\linewidth]{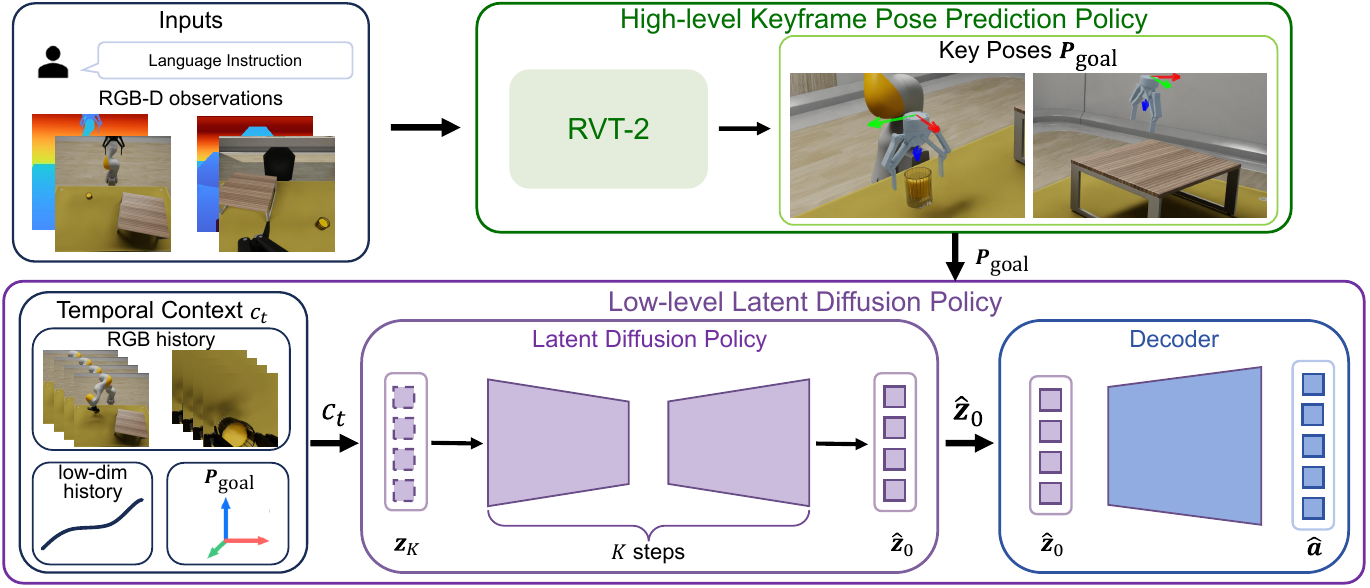}
    \caption{Architecture of the proposed hierarchical latent diffusion policy. The high-level policy predicts task-relevant SE(3) sub-goals $\mathbf{P}_{\mathrm{goal}}$ from language instructions and multi-view RGB-D observations. Conditioned on the predicted target pose, recent visual observations, and proprioceptive state history, the low-level latent diffusion policy iteratively denoises in the learned latent action space to generate a latent action sequence $\hat{\mathbf{z}}_0$. A lightweight decoder then maps $\hat{\mathbf{z}}_0$ into dense executable action chunks $\hat{\mathbf{a}}$ for smooth and high-frequency execution.}
    \label{fig:policy_framework}
    \vskip -15pt
\end{figure*}

We propose a hierarchical  control framework for stable and smooth pick-and-place tasks, as illustrated in \cref{fig:policy_framework}.
The key motivation is that dynamically sensitive manipulation involves heterogeneous requirements across temporal scales. 
Task-level reasoning determines where to grasp and place from language and multi-view observations, while execution-level control must generate high-frequency, temporally coherent actions without abrupt corrections. 
Coupling these requirements into a monolithic observation-to-action policy makes it difficult to balance low-frequency semantic reasoning with high-frequency smooth control.
Moreover, directly applying diffusion to dense action chunks in the original action space can be computationally expensive and may introduce temporally inconsistent low-level motions, which are undesirable for stability-sensitive execution.
We therefore decompose control into a high-level keyframe policy and a low-level latent diffusion policy. 
The high-level policy predicts $\text{SE}(3)$ sub-goals for the current manipulation stage.
Conditioned on each sub-goal, the low-level policy first generates compact latent action sequences through diffusion and then decodes them into dense executable action chunks for stable execution.

\subsection{High-level Keyframe Pose Prediction Policy}
% next-best end-effector pose predict

The high-level policy $\pi_H$ maps language instructions $l$ and multi-view RGB-D observations $\mathcal{O}_t$ to the next keyframe end-effector pose $\mathbf{P}_{\text{goal}} \in \text{SE}(3)$, which represents task-level sub-goals such as grasping or placing poses:
\begin{equation}
    \mathbf{P}_{\text{goal}} = \pi_{H}(\mathcal{O}_t,l).
\end{equation}
During data generation, we directly record the end-effector poses at the successful grasp and placement events as the two keyframe annotations, requiring neither manual labeling nor velocity-based post-processing. 
We employ RVT-2~\cite{rvt2} as the high-level policy and retain its original architecture. RVT-2's design features a virtual-view re-rendering mechanism that enhances camera-viewpoint generalization, alongside a multi-stage refinement pipeline that improves pose predictions.
The predicted sub-goals are passed to the low-level policy, decoupling semantic reasoning from smooth motion generation.

\subsection{Low-level Latent Diffusion Policy}
The primary objective of the low-level policy is to translate the sparse $\text{SE}(3)$ sub-goals predicted by the high-level policy into smooth, high-frequency end-effector trajectories. 
A straightforward solution is to employ the standard Diffusion Policy~\cite{diffusion_policy} to directly generate dense action chunks in the raw action space. 
However, its iterative denoising process is computationally costly for high-frequency control, and small inconsistencies across successive predicted chunks can lead to oscillatory motions, increasing the risk of fluid spillage.

Inspired by recent latent-space diffusion~\cite{hdp, graspldp}, we propose a Latent Diffusion Policy that separates efficient motion generation from dense trajectory reconstruction. 
Specifically, we train an action autoencoder that compresses a dense action chunk into a lower-dimensional latent action sequence along both the temporal and action-feature dimensions, and reconstructs it back into a dense, executable trajectory.
The diffusion policy generates a motion plan in this compact latent space, reducing inference latency compared to raw-action diffusion. 
Then, a lightweight decoder decodes the predicted latent representations into dense, high-frequency action chunks for execution. 
This design preserves the generative flexibility of diffusion models while improving both computational efficiency and temporal consistency for stable manipulation.

\noindent\textbf{Action Autoencoder.}
The action autoencoder is formulated as a variational autoencoder (VAE) that compresses dense action chunks along both the temporal and action-feature dimensions, constructing a compact latent space for downstream diffusion-based trajectory generation.
Given a dense action chunk $\mathbf{a} \in \mathbb{R}^{T \times D_a}$, the encoder $\mathcal{E}_{\psi}$ predicts the parameters of a diagonal Gaussian posterior:
\begin{equation}
    q_{\psi}(\mathbf{z} \mid \mathbf{a})
    =
    \mathcal{N}\big(
    \boldsymbol{\mu}_{\psi}(\mathbf{a}),
    \mathrm{diag}(\boldsymbol{\sigma}_{\psi}^{2}(\mathbf{a}))
    \big),
\end{equation}
where $\mathbf{z} \in \mathbb{R}^{T' \times D_z}$, $T' < T$, and $D_z < D_a$.
During training, a latent action sequence $\tilde{\mathbf{z}}$ is sampled from this posterior using the reparameterization trick and then reconstructed by a lightweight decoder:
\begin{equation}
\hat{\mathbf{a}} = \mathcal{D}_{\phi}(\tilde{\mathbf{z}}),
\quad
\hat{\mathbf{a}} \in \mathbb{R}^{T \times D_a},
\end{equation}
where $\mathcal{D}_{\phi}$ is implemented as a multilayer perceptron (MLP).
The encoder and decoder are trained jointly with reconstruction and KL regularization losses:
\begin{equation}
\begin{split}
\mathcal{L}_{\mathrm{AE}} =\;&
\lambda_{1}
\left\|
\mathbf{a} - \hat{\mathbf{a}}
\right\|_1
+
\lambda_{2}
\left\|
\mathbf{a} - \hat{\mathbf{a}}
\right\|_2^2 \\
&\quad
+
\lambda_{\mathrm{KL}}
D_{\mathrm{KL}}\big(
q_{\psi}(\mathbf{z} \mid \mathbf{a})
\,\|\,
\mathcal{N}(\mathbf{0}, \mathbf{I})
\big).
\end{split}
\label{eq:loss-fine}
\end{equation}
The reconstruction terms encourage accurate recovery of dense executable trajectories, while the KL term regularizes the latent posterior toward a standard Gaussian prior, yielding a continuous latent space for diffusion-based generation.

\noindent\textbf{Latent Diffusion Policy.}
We formulate action generation as a goal-conditioned diffusion process in a continuous latent action space.
After training the action autoencoder, each demonstration action chunk $\mathbf{a}$ is encoded into a deterministic latent action representation as the clean target for diffusion by taking the posterior mean of the encoder:
\begin{equation}
\mathbf{z}_0 = \boldsymbol{\mu}_{\psi}(\mathbf{a}).
\label{eq:latent-mean}
\end{equation}
To guide the generation process of $\mathbf{z}_0$, the policy is conditioned on a multi-modal temporal context $\mathbf{c}$ that integrates visual perception, proprioceptive state history, and goal specification. Formally, the conditioning signal is defined as $\mathbf{c} = \{\mathbf{o}_{t-h:t}^{\text{rgb}}, \mathbf{o}_{t-l:t}^{\text{prop}}, \mathbf{P}_{\text{goal}}\}$, where:
(i) $\mathbf{o}_{t-h:t}^{\text{rgb}}$ denotes a short-horizon sequence of RGB observations that provides recent visual feedback from the scene;
(ii) $\mathbf{o}_{t-l:t}^{\text{prop}}$ represents a longer history of robot proprioceptive states, capturing past motion dynamics and trajectory evolution; and
(iii) $\mathbf{P}_{\text{goal}} \in \text{SE}(3)$ specifies the target end-effector pose predicted by the high-level policy.
Together, these conditioning inputs enable the policy to adapt to short-term visual changes, maintain temporal consistency with the motion history, and generate actions directed toward the target pose.

We model the distribution of latent action sequences with a conditional Denoising Diffusion Probabilistic Model (DDPM)~\cite{ddpm}.
During training, we perturb the clean latent action sequence $\mathbf{z}_0$ defined in \eqref{eq:latent-mean} with Gaussian noise at diffusion step $k$:
\begin{equation} 
    \mathbf{z}_k = \sqrt{\bar{\alpha}_k}\mathbf{z}_0 
    + 
    \sqrt{1-\bar{\alpha}_k}\boldsymbol{\epsilon}, 
    \quad 
    \boldsymbol{\epsilon} \sim \mathcal{N}(\mathbf{0}, \mathbf{I}), 
\end{equation} 
where $\bar{\alpha}_k$ denotes the cumulative noise schedule.
A conditional noise prediction network $\boldsymbol{\epsilon}_{\theta}$, implemented as a 1D U-Net, takes the noisy latent sequence $\mathbf{z}_k$, the conditioning context $\mathbf{c}$, and the diffusion step $k$ as input, and predicts the injected noise. 
The latent diffusion policy is trained with the standard noise prediction objective: \begin{equation} \mathcal{L}_{\mathrm{LDP}} = 
\mathbb{E}_{\mathbf{z}_0, \mathbf{c}, \boldsymbol{\epsilon}, k} 
\left[ 
\left\| 
\boldsymbol{\epsilon} 
- 
\boldsymbol{\epsilon}_{\theta}(\mathbf{z}_k, \mathbf{c}, k) 
\right\|_2^2 
\right]. 
\label{eq:loss-coarse}
\end{equation}

During inference, we sample an initial latent noise sequence $\mathbf{z}_K \sim \mathcal{N}(\mathbf{0}, \mathbf{I})$ and iteratively denoise it conditioned on $\mathbf{c}$ for $K$ steps to generate the predicted latent action sequence $\hat{\mathbf{z}}_0$. 
The resulting latent sequence is then fed into the decoder $\mathcal{D}_{\phi}$  to generate dense executable action chunks.
Following ACT~\cite{act}, we apply temporal ensemble across consecutive decoded chunks, where overlapping predictions for the same execution timestep are aggregated with exponentially decayed weights. 
This reduces chunk-boundary discontinuities and improves execution smoothness.

\section{EXPERIMENTS}
\begin{figure*}[t]
    \centering
    \includegraphics[width=\linewidth]{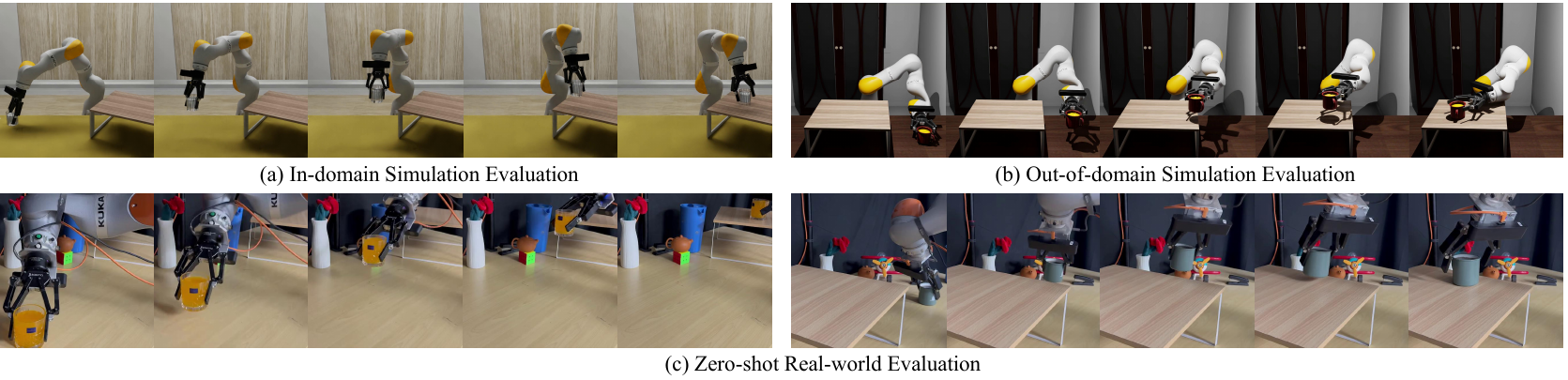}
    \caption{
Qualitative results in simulation and the real world.
(a) In-domain simulation evaluation on seen distributions.
(b) Out-of-domain simulation evaluation on unseen objects and scene configurations.
(c) Zero-shot real-world evaluation, demonstrating the sim-to-real transfer of the learned policy.
}
    \label{fig:exp_vis}
    \vskip -15pt
\end{figure*}

\subsection{Experimental Setup}
\noindent\textbf{Benchmark.}
We evaluate stable manipulation on a liquid transport benchmark, where a robot is required to grasp a liquid-filled container, transport it to a designated target platform, and place it successfully while suppressing sloshing and preventing spillage.
We construct the benchmark using the synthetic data generation pipeline described in \cref{sim}. 
We use diverse cup assets from Isaac Sim and Objaverse~\cite{objaverse}, covering different materials, shapes, and visual appearances, together with randomized tabletop textures and scene settings. 
We introduce a raised target platform to increase spatial difficulty and evaluate stable transport across height variations. 
In total, we generate 12 scene variations with 100 demonstrations each, resulting in 1200 training episodes.

To evaluate both in-distribution performance and out-of-domain (OOD) robustness, we test all methods on 50 configurations for each of four distinct splits: 
(1) \textbf{In-domain}, following the training distribution; 
(2) \textbf{OOD-Object}, utilizing unseen cup instances; 
(3) \textbf{OOD-Scene}, using unseen tabletop textures and backgrounds; and 
(4) \textbf{OOD-Object+Scene}, combining both shifts to assess compound generalization. 
For a fair comparison, all evaluations use the same fixed initializations and are reported over three random seeds.

\noindent\textbf{Implementation Details.}
The high-level RVT-2 policy is trained on the generated keyframe annotations using the original training recipe.
The low-level policy is trained sequentially in two stages.

\noindent\textbf{Stage 1: Action Autoencoder.}
A CNN encoder compresses a $32 \times 10$ action chunk into an $8 \times 8$ latent action representation, which an MLP decoder reconstructs into raw actions (3D position, 6D rotation, and 1D gripper width).
We optimize this model using the autoencoder loss $\mathcal{L}_{\text{AE}}$ defined in \eqref{eq:loss-fine} with AdamW for 600 epochs with a batch size of 64, a learning rate of $1 \times 10^{-3}$, and a weight decay of $1 \times 10^{-4}$.

\noindent\textbf{Stage 2: Latent Diffusion Policy.}
The policy predicts these $8 \times 8$ latent action representations conditioned on four visual steps, a 20-step proprioceptive history, and the target keyframe pose. It uses a ResNet-18 image encoder and a 100-step DDIM scheduler. 
Optimization is performed using the latent diffusion loss $\mathcal{L}_{\text{LDP}}$ defined in \eqref{eq:loss-coarse} with AdamW for 400 epochs with a batch size of 64, a learning rate of $1 \times 10^{-4}$, a weight decay of $1 \times 10^{-6}$, EMA, and a cosine decay schedule.

\noindent\textbf{Inference efficiency.} 
On an RTX 3090, LDP uses 300M parameters, 93.94\,ms per inference, and 1.19\,GB peak GPU memory, versus 330M, 107.06\,ms, and 1.33\,GB for raw-action DP, reducing latency by 12.3\% and memory usage by 10.5\%.

\noindent\textbf{Baselines.}
We compare our method against three representative baselines. 
First, \textbf{RVT-2 + DP} is a hierarchical baseline that shares our high-level module but replaces the low-level latent diffusion with a standard Diffusion Policy~\cite{diffusion_policy} operating in the raw action space. This baseline isolates the impact of our latent-space action generation design while keeping the hierarchical structure identical.
Next, \textbf{$\pi_{0.5}$ (few-shot)} evaluates the pre-trained generalist VLA model $\pi_{0.5}$~\cite{pi05} fine-tuned on 10\% of our training data, assessing whether general VLA priors can rapidly adapt to dynamically sensitive manipulation. 
Finally, \textbf{$\pi_{0.5}$ (full data)} fine-tunes the same VLA model on the complete training dataset. This setting evaluates whether a monolithic VLA model can achieve comparable stability and smoothness when trained with the same task-specific demonstrations.

\noindent\textbf{Metrics.}
To evaluate stable and smooth manipulation, we use four metrics that jointly measure task completion, object stability, and motion smoothness. 
To ensure meaningful and fair comparisons, the latter three metrics (OE, MTA, and TS) are computed exclusively over successful trials.

\textbf{Success Rate} (SR, \%) measures the percentage of trials in which the robot successfully grasps the cup, transports it to the target platform, and maintains the fluid loss below 5\% of the total fluid particles. 

\textbf{Orientation Error} (OE, $\mathrm{rad}\cdot\mathrm{s}$) evaluates the cumulative cup deviation from the upright direction during execution:
\begin{equation}
E_{\mathrm{orient}} = \int_{0}^{T} \theta_{\mathrm{tilt}}(t) \, \mathrm{d}t,
\end{equation}
where $\theta_{\mathrm{tilt}}(t)$ is the angle between the cup's central axis and the global vertical axis at time $t$, and $T$ is the task duration. 
Lower OE indicates more consistent maintenance of the upright pose throughout transport.

\textbf{Maximum Tilt Angle} (MTA, $^{\circ}$) measures the worst-case cup deviation during execution:
\begin{equation}
\theta_{\max} = \max_{t \in [0,T]} \theta_{\mathrm{tilt}}(t).
\end{equation}
This metric captures transient instability that may be averaged out or missed by the cumulative orientation error. 

\textbf{Trajectory Smoothness} (TS, $\mathrm{m}^2/\mathrm{s}^6$) measures the average squared jerk of the end-effector trajectory:
\begin{equation}
J = \frac{1}{T}\int_{0}^{T} \left\| \dddot{\mathbf{x}}(t) \right\|_2^2 \, \mathrm{d}t,
\end{equation}
where $\dddot{\mathbf{x}}(t)$ denotes the end-effector jerk. Lower TS indicates smoother motion with fewer abrupt acceleration changes, which is critical for reducing liquid sloshing.

\subsection{Simulation Evaluation}
We first evaluate the proposed framework against the baselines in simulation, assessing both in-domain performance and out-of-domain robustness across four evaluation splits.

\begin{table}[t]
    \centering
    \caption{In-Domain and Out-of-Domain Simulation Evaluation}
    \label{tab:sim}

    \footnotesize
    \setlength{\tabcolsep}{3pt}

    \begin{tabular}{lcccc}
        \toprule
        Method
        & SR $\uparrow$
        & OE $\downarrow$
        & MTA $\downarrow$
        & TS $\downarrow$ \\
        \midrule

        \multicolumn{5}{c}{\textbf{In-Domain Evaluation}} \\
        \midrule

        $\pi_{0.5}$ (few-shot)
        & \meanstd{11.13}{0.52}
        & \meanstd{1.33}{0.33}
        & \meanstd{18.53}{4.22}
        & \meanstd{277.29}{115.65} \\

        $\pi_{0.5}$ (full data)
        & \meanstd{49.03}{4.75}
        & \meanstd{1.23}{0.10}
        & \meanstd{17.29}{0.35}
        & \meanstd{272.43}{20.24} \\

        RVT-2 + DP
        & \secondmeanstd{69.74}{9.71}
        & \secondmeanstd{0.62}{0.06}
        & \secondmeanstd{9.71}{0.63}
        & \secondmeanstd{80.19}{6.48} \\

        RVT-2 + LDP (Ours)
        & \bestmeanstd{80.23}{0.55}
        & \bestmeanstd{0.55}{0.03}
        & \bestmeanstd{8.87}{0.29}
        & \bestmeanstd{61.98}{4.10} \\

        \midrule
        \multicolumn{5}{c}{
            \textbf{Out-of-Domain Evaluation (Object)}
        } \\
        \midrule

        $\pi_{0.5}$ (few-shot)
        & \meanstd{5.33}{4.62}
        & \meanstd{0.96}{0.04}
        & \meanstd{16.51}{1.12}
        & \meanstd{155.45}{10.76} \\

        $\pi_{0.5}$ (full data)
        & \meanstd{41.60}{1.46}
        & \meanstd{0.91}{0.07}
        & \meanstd{15.34}{0.62}
        & \meanstd{207.92}{57.91} \\

        RVT-2 + DP
        & \secondmeanstd{69.45}{7.60}
        & \secondmeanstd{0.80}{0.09}
        & \secondmeanstd{11.52}{1.07}
        & \secondmeanstd{83.02}{8.89} \\

        RVT-2 + LDP (Ours)
        & \bestmeanstd{76.42}{0.29}
        & \bestmeanstd{0.67}{0.02}
        & \bestmeanstd{9.57}{0.11}
        & \bestmeanstd{68.35}{5.78} \\

        \midrule
        \multicolumn{5}{c}{
            \textbf{Out-of-Domain Evaluation (Scene)}
        } \\
        \midrule

        $\pi_{0.5}$ (few-shot)
        & \meanstd{2.97}{2.57}
        & \meanstd{2.02}{0.69}
        & \meanstd{23.12}{2.57}
        & \meanstd{226.48}{40.72} \\

        $\pi_{0.5}$ (full data)
        & \meanstd{30.22}{6.40}
        & \meanstd{1.28}{0.17}
        & \meanstd{18.83}{0.99}
        & \meanstd{178.30}{21.34} \\

        RVT-2 + DP
        & \secondmeanstd{71.79}{8.58}
        & \secondmeanstd{0.75}{0.14}
        & \secondmeanstd{9.48}{0.77}
        & \secondmeanstd{86.23}{12.98} \\

        RVT-2 + LDP (Ours)
        & \bestmeanstd{73.30}{4.41}
        & \bestmeanstd{0.61}{0.05}
        & \bestmeanstd{8.34}{0.38}
        & \bestmeanstd{65.13}{5.99} \\

        \midrule
        \multicolumn{5}{c}{
            \textbf{Out-of-Domain Evaluation (Object + Scene)}
        } \\
        \midrule

        $\pi_{0.5}$ (few-shot)
        & \meanstd{9.94}{1.21}
        & \bestmeanstd{0.61}{0.48}
        & \meanstd{11.92}{5.48}
        & \meanstd{191.61}{5.07} \\

        $\pi_{0.5}$ (full data)
        & \meanstd{27.58}{7.74}
        & \meanstd{1.03}{0.11}
        & \meanstd{16.07}{0.82}
        & \meanstd{147.21}{9.42} \\

        RVT-2 + DP
        & \secondmeanstd{48.42}{1.56}
        & \meanstd{0.74}{0.07}
        & \secondmeanstd{10.04}{0.84}
        & \secondmeanstd{85.03}{15.14} \\

        RVT-2 + LDP (Ours)
        & \bestmeanstd{70.08}{10.83}
        & \secondmeanstd{0.63}{0.11}
        & \bestmeanstd{8.58}{1.03}
        & \bestmeanstd{58.13}{5.01} \\

        \bottomrule
    \end{tabular}
    \vskip -15pt
\end{table}

\noindent\textbf{In-domain Evaluation.} 
As shown in \cref{tab:sim}, our method achieves the best overall performance among all methods. 
Compared with RVT-2 + DP, which shares the same high-level module but generates actions in the raw action space, our method improves the success rate by 10.49\%. 
It also achieves lower OE, MTA, and TS, indicating better upright-pose maintenance and reduced transient instability during transport. 
These results show that generating low-level motions in a compact latent space is more effective than directly diffusing dense raw actions for stable manipulation.
The VLA baselines perform substantially worse in this dynamically sensitive setting. 
Although full-data fine-tuning improves $\pi_{0.5}$ over the few-shot variant, its success rate remains far below that of the hierarchical methods.
This indicates that general VLA priors alone are insufficient for stable manipulation, and that directly learning end-to-end actions remains challenging even with task-specific demonstrations. 
Qualitative rollouts of our policy, provided in the attached video and \cref{fig:exp_vis}(a), demonstrate its ability to transport the container with minimal sloshing while maintaining a consistent upright pose throughout execution.

\noindent\textbf{Out-of-domain Evaluation.} 
As shown in \cref{tab:sim}, our method consistently achieves the highest success rate across all OOD splits, indicating strong generalization of the learned stability-aware behavior.
This advantage becomes most pronounced in the challenging OOD-object+scene setting, where our method improves upon {RVT-2 + DP} by 21.66\% in success rate.
The stability metrics also show consistent gains. 
Compared with RVT-2 + DP, our method generally achieves lower OE, MTA, and TS, suggesting that latent-space action generation provides more robust low-level control under distribution shifts. 
In contrast, the VLA baselines exhibit limited OOD robustness, especially under the compound shift, indicating that general VLA priors alone are insufficient for stable manipulation. 
Qualitative results of our policy in the attached video and \cref{fig:exp_vis}(b) illustrate its robust performance under distribution shifts, showcasing stable transport with minimized tilt and smooth motion.

\subsection{Real-world Evaluation}
\begin{table}[t]
    \centering
    \caption{Real-world Evaluation}
    \setlength{\tabcolsep}{3.5pt}
    \begin{tabular}{lcccc}
        \toprule
        Method & SR $\uparrow$ & OE $\downarrow$ & MTA $\downarrow$ & TS $\downarrow$ \\
        \midrule
        $\pi_{0.5}$ (full data) & 36.67 & 0.98 & 10.56 & 150.58  \\
        RVT-2 + DP & \underline{50.00} & \underline{0.70} & \underline{5.88} & \underline{63.46} \\
        RVT-2 + LDP (Ours) & \textbf{73.33} & \textbf{0.65} & \textbf{3.96} & \textbf{50.73} \\
        \bottomrule
    \end{tabular}
    \label{tab:real}
    \vskip -15pt
\end{table}

We further evaluate our framework on a physical robotic platform to examine its zero-shot sim-to-real transfer capability.
The platform consists of a KUKA LBR iiwa 14 R820 robot arm equipped with a Robotiq 2F-140 gripper. 
For perception, we utilize two ZED 2i RGB-D cameras: a front camera observing the workspace and a wrist camera.

During real-world deployment, we adopt a latency-aware asynchronous execution strategy~\cite{dynamicvla} to avoid inference-induced pauses, which are particularly harmful for liquid transport.
While the robot executes the current action chunk, the policy infers the next chunk from the latest observations.
Once the new chunk is available, actions corresponding to timesteps already elapsed during inference are discarded, and the remaining future actions are temporally ensembled with overlapping predictions for smooth continuous execution.

We evaluate the policies on a diverse set of real-world cups made of glass, plastic, and metal, which vary significantly in shape, size, reflectance, and visual appearance. 
Each policy is evaluated over 30 real-world trials. 
We omit \textbf{$\pi_{0.5}$ (few-shot)} from the real-world evaluation because its preliminary rollouts were highly unstable and unsafe for liquid transport.
Crucially, all reported policies are trained purely on synthetic demonstrations, without any real-world demonstrations, fine-tuning, or adaptation. 
Thus, the real-world evaluation serves as a direct test of both the quality of our stability-aware synthetic data and the sim-to-real generalization of the learned policy. 
Qualitative real-world rollouts of our policy are provided in the attached video and \cref{fig:exp_vis}(c).

As shown in \cref{tab:real}, our method achieves the best overall real-world performance among all compared methods. 
Since all policies are deployed without real-world adaptation, the results indicate that our simulation pipeline generates demonstrations that capture transferable stability-aware behaviors. 
Compared with RVT-2 + DP, our method achieves better task success, stability, and smoothness, demonstrating the advantage of the latent-space action generation over raw-action diffusion for high-frequency real-world execution. 
The gap over $\pi_{0.5}$ further suggests that dynamically sensitive manipulation requires more than generalist semantic and visual priors; it demands an architecture capable of generating temporally coherent, stability-aware actions. 
Overall, the real-world results demonstrate the zero-shot sim-to-real capability of our hierarchical policy and validate the effectiveness of training purely on synthetic, stability-aware demonstrations.

\subsection{Ablation Study}
\begin{table}[t]
    \centering
    \footnotesize
    \caption{Ablation Studies}
    \label{tab:ablation}
    \setlength{\tabcolsep}{2pt}
    \begin{tabular}{llcccc}
        \toprule
        Component & Variant & SR $\uparrow$ & OE $\downarrow$ & MTA $\downarrow$ & TS $\downarrow$ \\
        \midrule
        Full model 
        & --
        & \bestmeanstd{70.08}{10.83}
        & \secondmeanstd{0.63}{0.11}
        & \bestmeanstd{8.58}{1.03}
        & \bestmeanstd{58.13}{5.01} \\
        \midrule
        Data pipeline 
        & w/o Traj. Filter 
        & \meanstd{54.02}{11.08}
        & \meanstd{0.72}{0.03}
        & \meanstd{10.07}{0.34}
        & \meanstd{59.42}{6.47} \\
        
        Hierarchy 
        & LDP-only 
        & \meanstd{29.71}{9.99}
        & \meanstd{0.89}{0.08}
        & \meanstd{10.77}{1.23}
        & \meanstd{158.89}{38.10} \\
        
        % History 
        \multirow{2}{*}{History}
        & Hist-4 
        & \meanstd{55.48}{3.91}
        & \meanstd{0.67}{0.15}
        & \meanstd{9.14}{0.92}
        & \meanstd{64.14}{3.94} \\
        & Hist-12 
        & \secondmeanstd{57.40}{11.45}
        & \bestmeanstd{0.61}{0.04}
        & \secondmeanstd{8.90}{0.77}
        & \secondmeanstd{58.78}{5.27} \\
        
        Aggregation 
        & w/o Ensemble 
        & \meanstd{18.37}{1.06}
        & \meanstd{2.11}{0.37}
        & \meanstd{16.38}{1.28}
        & \meanstd{218.47}{47.71} \\
        \bottomrule
    \end{tabular}
    \vskip -15pt
\end{table}

We evaluate the contribution of each key component under the OOD-object+scene setting, as reported in \cref{tab:ablation}.

\noindent\textbf{Stability-aware data filtering.} 
Removing both tilt screening and fluid simulation (\textbf{w/o Traj. Filter}) degrades both success rate and stability.
This indicates that geometric and kinematic feasibility alone is insufficient for stable dynamics, underscoring the necessity of trajectory-level dynamic validation.

\noindent\textbf{Hierarchical keyframe prediction.}
Removing the high-level keyframe policy (\textbf{LDP-only}) causes a substantial performance drop. This confirms that the low-level policy alone struggles to infer both task-level goals and smooth executable trajectories under compound shifts, justifying the decomposition of semantic reasoning and motion generation.

\noindent\textbf{Proprioceptive history length.}
Shortening proprioceptive history (\textbf{Hist-4}, \textbf{Hist-12}) reduces success rates compared to our full \textbf{Hist-20} model. While \textbf{Hist-12} maintains comparable stability, \textbf{Hist-20} achieves the strongest overall performance, demonstrating that longer history provides crucial temporal context for execution under distribution shifts.

\noindent\textbf{Temporal ensemble.}
Removing temporal ensembling (\textbf{w/o Ensemble}) causes the most significant performance degradation. This component is critical for stable closed-loop execution as it suppresses abrupt chunk-boundary transitions, thereby reducing spill-inducing motions under unseen configurations.

\section{CONCLUSION}
In this work, we present a synthetic-to-real framework for the dynamically stable and smooth manipulation of liquid-filled containers. 
Our approach combines a stability-aware synthetic data pipeline with a hierarchical latent diffusion policy, enabling the generation of smooth trajectories. 
Evaluations show that our framework improves success rates, trajectory stability, and motion smoothness in both simulation and real world compared to representative baselines. 
Notably, real-world experiments demonstrate zero-shot sim-to-real generalization without any real-world demonstrations. 
Future work will extend this framework to other dynamically sensitive manipulation tasks beyond liquid transport.

\bibliographystyle{IEEEtran}
\bibliography{references}

\newpage

\vfill

\end{document}